\documentclass[11pt]{article}

\usepackage[utf8]{inputenc}
\usepackage[T1]{fontenc}
\usepackage{mathptmx}
\usepackage[margin=1in]{geometry}
\usepackage{amsmath,amssymb}
\usepackage{graphicx}
\usepackage{booktabs}
\usepackage{array}
\usepackage{caption}
\usepackage{xcolor}
\usepackage{tikz}
\usetikzlibrary{arrows.meta,positioning,shapes.geometric,calc,fit,backgrounds,shadows}
\usepackage{listings}
\usepackage{enumitem}
\usepackage{authblk}
\usepackage[hidelinks]{hyperref}
\usepackage{microtype}

\setlist{itemsep=2pt,topsep=3pt,parsep=0pt}

\definecolor{navy}{HTML}{1F3A5F}
\definecolor{rust}{HTML}{B5462A}
\definecolor{teal}{HTML}{2F7E7E}
\definecolor{green}{HTML}{2E7D32}
\definecolor{lgrey}{HTML}{F4F4F4}
\definecolor{codegrey}{HTML}{F7F7F7}
\definecolor{codeframe}{HTML}{CCCCCC}
\definecolor{keywordblue}{HTML}{0000AA}
\definecolor{stringgreen}{HTML}{007700}
\definecolor{commentgrey}{HTML}{777777}

\lstdefinelanguage{JavaScript}{
  keywords={const,let,var,function,return,if,else,await,async,new,true,false,null},
  keywordstyle=\color{keywordblue}\bfseries,
  comment=[l]{//},
  morecomment=[s]{/*}{*/},
  commentstyle=\color{commentgrey}\itshape,
  stringstyle=\color{stringgreen},
  morestring=[b]",
  morestring=[b]',
  morestring=[b]`,
  sensitive=true
}
\title{\vspace{-1.2cm}\bfseries A Mathematical Forum Platform for Collaborative\\
Problem Solving and Dataset Generation for AI Reasoning}

\author[1]{Nurmukhammad Abdurasulov}
\author[1]{Akbar Erkinov}
\affil[1]{\small Independent Researchers, San Francisco, CA, USA}
\date{}

\begin{document}
\maketitle
\vspace{-0.8cm}

\begin{abstract}
\noindent
Sharing mathematical content in online forums remains a significant friction point for students
and educators: writing raw \LaTeX{} is error-prone, standalone optical-character-recognition
(OCR) tools require platform switching, and current forum software offers no integrated path
from a photograph of a formula to a rendered post. We present a unified system that eliminates
this friction by embedding an image-to-\LaTeX{} conversion pipeline directly inside a forum
posting interface. A user uploads or captures an image of a mathematical expression; the system
routes it through the Mathpix OCR API, detects whether the returned output is \LaTeX{} or
plain text containing inline math, applies the appropriate delimiter normalisation, and renders
a live preview in either \LaTeX{} or Markdown mode before the post is committed to the
database. The architecture is organized in three loosely coupled layers---image processing,
rendering, and storage---and supports both desktop and mobile clients. A provisional US patent
application (No.~63/727,195) has been filed covering the core methods. We describe the full
system design, each component in detail, the data schema, and the key technical innovations,
and we position the work against existing standalone tools and forum platforms to demonstrate
the practical gap it closes. Beyond immediate usability, we argue that a deployed platform
of this kind constitutes a continuously growing, community-validated dataset of mathematical
problems and step-by-step solutions---a resource that can be used to train and benchmark AI
systems for accurate mathematical reasoning.
\end{abstract}


\section{Introduction}
Online forums have become the de facto venue for collaborative mathematics: platforms such as
Mathematics Stack Exchange, Piazza, and course-embedded discussion boards handle millions of
questions involving formulas, proofs, and equations every year. Yet the experience of
\emph{posting} mathematics remains awkward. The dominant entry method---hand-writing raw
\LaTeX{} source---requires fluency in a typesetting language that many students never acquire
confidently~\cite{lamport1994}. Errors are common, previews are not always available,
and the feedback loop between typing and seeing a rendered result can be slow. Alternative
approaches exist but each introduces its own friction: a student who photographs a homework
problem must open a separate OCR application (such as Mathpix Snip~\cite{mathpix} or
pix2tex~\cite{pix2tex}), copy the returned \LaTeX{} string, switch back to the forum
interface, paste, adjust delimiters, and then preview. Each context switch is an opportunity
for error and a disincentive to engage.

\begin{figure}[t]
\centering
\includegraphics[width=0.78\linewidth]{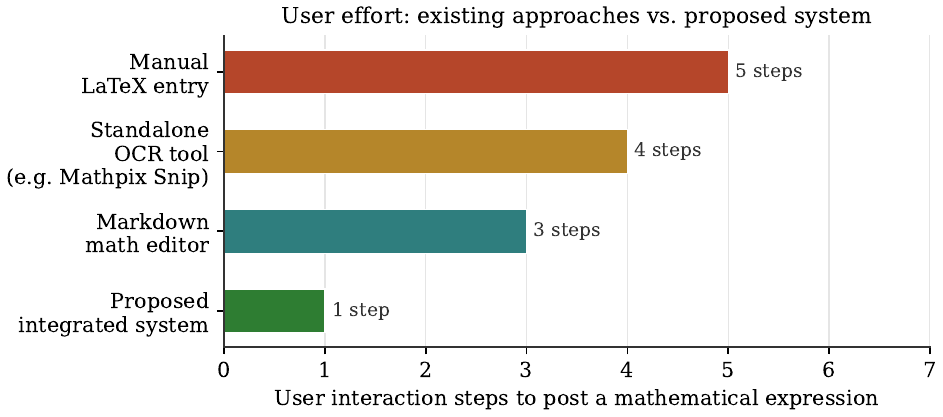}
\caption{User interaction steps required to post a mathematical expression using four
approaches. The proposed integrated system reduces the workflow to a single step: upload
an image within the forum interface.}
\label{fig:steps}
\end{figure}

The root cause is that the three operations a student needs---image capture, formula
recognition, and forum posting---live in separate applications with no shared state. The
contribution of this paper is a system that eliminates the seams between them: \emph{image
capture, OCR processing, format detection, real-time preview, and forum posting operate as a
single, uninterrupted workflow}. Figure~\ref{fig:steps} illustrates the reduction in user
interaction steps this achieves compared to existing approaches.

The system builds on the Mathpix OCR API~\cite{mathpix}, which has demonstrated strong
performance on both printed and handwritten mathematical content~\cite{pdf2latex}, and on
MathJax~\cite{mathjax} for client-side rendering. The key contributions are not the underlying
OCR or rendering engines, which are established, but rather: (i) the integration architecture
that connects them seamlessly within a forum interface; (ii) the format-handler component that
detects, normalises, and routes conversion output to the appropriate renderer without user
intervention; and (iii) a dual-format rendering system that preserves author intent while
accommodating both pure-\LaTeX{} and Markdown-with-math posts. A provisional patent application
(US~63/727,195, filed December~3, 2024) covers the core methods.

Beyond its immediate usability benefits, the platform has a second, longer-term value: every
problem posted and every solution contributed becomes a structured, community-validated data
record. At scale, this accumulation of problem--solution pairs---each anchored by an image,
a machine-readable \LaTeX{} representation, and a natural-language or symbolic solution---
constitutes a rich dataset for training AI systems to solve mathematics accurately. The
scarcity of large, high-quality, verified mathematical datasets is a well-known bottleneck in
this field~\cite{gsm8k,math_dataset}, and a forum that lowers the barrier to posting
mathematics naturally produces such data as a by-product of normal community activity.

The remainder of the paper is organised as follows. Section~\ref{sec:related} reviews related
work. Section~\ref{sec:overview} gives a system overview. Sections~\ref{sec:arch}--\ref{sec:impl}
detail the architecture and implementation. Section~\ref{sec:innovations} discusses the key
innovations. Section~\ref{sec:dataset} describes the platform's role as a source of
mathematical training data for AI. Sections~\ref{sec:limits}--\ref{sec:conclusion} address
limitations, future work, and conclusions.

\section{Related Work}
\label{sec:related}

\subsection{Mathematical expression recognition}
Recognition of mathematical notation from images is a long-standing research problem. Early
work by Anderson~\cite{anderson1967} established the linguistic analogy between mathematical
layout and formal grammars, and subsequent decades produced increasingly capable systems using
specialized character segmentation combined with grammatical models of mathematical
structure~\cite{miller1998,chan2000}. The INFTY system~\cite{infty2003} exemplified this
generation: it converts scanned mathematical documents to \LaTeX{} and other markup formats
using a pipeline of symbol classification and structural analysis. A thorough survey of both
recognition and retrieval of mathematical expressions is given by Zanibbi and
Blostein~\cite{zanibbi2012}.

The advent of deep learning substantially changed the approach. Deng \emph{et al.}~\cite{deng2017}
introduced an attention-based encoder--decoder architecture---the im2latex model---that treats
the problem as sequence-to-sequence learning: a convolutional encoder maps the image to a
feature grid and a recurrent decoder with coarse-to-fine attention generates the \LaTeX{}
token sequence. Trained on the im2latex-100K dataset derived from arXiv papers, the model
outperformed classical OCR systems by a large margin. Blecher's pix2tex~\cite{pix2tex} extends
this direction with a Vision Transformer (ViT) encoder, achieving strong performance on
formulas encountered in the wild rather than in typeset documents. Commercial systems such as
Mathpix~\cite{mathpix} combine these techniques with large proprietary training sets and
production-grade APIs; independent evaluation has found Mathpix to be the most effective OCR
tool for \LaTeX{} conversion by BLEU and edit-distance metrics~\cite{pdf2latex}.

\subsection{Web-based mathematical rendering}
Displaying \LaTeX{} inside a browser requires a dedicated rendering engine because HTML has no
native support for mathematical typesetting at arbitrary complexity. MathJax~\cite{mathjax},
released in 2010 by the American Mathematical Society, the Society for Industrial and Applied
Mathematics, and Design Science, has become the gold standard: it processes \LaTeX{} or MathML
embedded in a page and renders it to HTML with CSS or to SVG, entirely in JavaScript without
server-side involvement. KaTeX~\cite{katex}, developed by Khan Academy, trades MathJax's
completeness for significantly faster rendering, making it attractive for real-time preview
scenarios. Both engines accept the standard \texttt{\$...\$} and \texttt{\$\$...\$\$}
delimiters for inline and display math respectively, but differ in their handling of the
alternative \texttt{\textbackslash(} \ldots \texttt{\textbackslash)} and
\texttt{\textbackslash[} \ldots \texttt{\textbackslash]} forms that the Mathpix API returns by
default in plain-text mode.

\subsection{Mathematical content in online forums}
Mathematics Stack Exchange, launched in 2010, was among the first major platforms to enable
MathJax rendering in user-submitted content~\cite{mathjax}. Users enter raw \LaTeX{}, and the
forum software passes it to MathJax for display. Discourse~\cite{discourse}, a widely deployed
open-source forum platform, offers a MathJax plugin but similarly requires users to enter
\LaTeX{} manually. Piazza, used by many university courses, supports \LaTeX{} in posts but
provides no image-capture pathway. Learning management systems such as Moodle and Canvas
provide formula editors but these are separate modal dialogs with no image input. To our
knowledge, no widely deployed forum system provides a native, end-to-end, image-to-rendered-post
workflow, which is the gap this system closes.

\section{System Overview}
\label{sec:overview}
\begin{figure}[t]
\centering
\begin{tikzpicture}[
  font=\footnotesize,
  node distance=5mm and 10mm,
  layer/.style={draw=navy!40, fill=navy!4, rounded corners=4pt,
                inner sep=5pt, font=\scriptsize\itshape\color{navy!70}},
  comp/.style={draw=navy, fill=navy!8, rounded corners=3pt,
               align=center, minimum height=8mm, inner sep=4pt,
               text width=22mm, font=\footnotesize},
  dec/.style={draw=navy, fill=navy!12, diamond, aspect=2, align=center,
              minimum height=8mm, inner sep=2pt, font=\scriptsize},
  io/.style={draw=teal, fill=teal!10, rounded corners=3pt,
             align=center, minimum height=8mm, inner sep=4pt,
             text width=22mm, font=\footnotesize},
  store/.style={draw=green!60!black, fill=green!6, rounded corners=3pt,
                align=center, minimum height=8mm, inner sep=4pt,
                text width=22mm, font=\footnotesize},
  arr/.style={-{Stealth[length=2mm]}, navy, thick},
  arrd/.style={-{Stealth[length=2mm]}, teal, thick, dashed},
]

\node[io] (ui) {User Interface};

\node[comp, below=10mm of ui] (upload) {Image Upload Component};
\node[dec, below=6mm of upload] (imtype) {Image Type};
\node[comp, below=8mm of imtype, xshift=-22mm] (mathpix) {Mathpix API Integration};
\node[comp, below=8mm of imtype, xshift=+22mm] (storage_api) {Storage API};

\begin{scope}[on background layer]
  \node[layer, fit=(upload)(imtype)(mathpix)(storage_api),
        label={[font=\scriptsize\itshape\color{navy!60}, anchor=north west]
               north west:{Image Processing Pipeline}}] (pipe_box) {};
\end{scope}

\node[comp, below=16mm of mathpix, xshift=22mm] (fmt) {Format Handler};
\node[comp, below=8mm of fmt, xshift=-18mm] (latex_r) {LaTeX Renderer};
\node[comp, below=8mm of fmt, xshift=+18mm] (md_r) {Markdown Math Renderer};

\begin{scope}[on background layer]
  \node[layer, fit=(fmt)(latex_r)(md_r),
        label={[font=\scriptsize\itshape\color{navy!60}, anchor=north west]
               north west:{Rendering System}}] (render_box) {};
\end{scope}

\node[store, below=14mm of fmt] (post) {Forum Post};
\node[store, below=6mm of post] (db) {Database};

\begin{scope}[on background layer]
  \node[layer, fit=(post)(db),
        label={[font=\scriptsize\itshape\color{green!50!black!70}, anchor=north west]
               north west:{Storage Layer}}] (store_box) {};
\end{scope}

\draw[arr] (ui) -- (upload);
\draw[arr] (upload) -- (imtype);
\draw[arr] (imtype) -- node[left, font=\scriptsize, xshift=-1mm]{Math} (mathpix);
\draw[arr] (imtype) -- node[right, font=\scriptsize, xshift=1mm]{Thumbnail} (storage_api);
\draw[arr] (mathpix) -- (fmt);
\draw[arr] (storage_api) |- (fmt);
\draw[arr] (fmt) -- node[left, font=\scriptsize]{\LaTeX} (latex_r);
\draw[arr] (fmt) -- node[right, font=\scriptsize]{Markdown} (md_r);
\draw[arr] (latex_r) |- (post);
\draw[arr] (md_r) |- (post);
\draw[arr] (post) -- (db);
\end{tikzpicture}
\caption{End-to-end system architecture. The three shaded regions correspond to the three
layers: Image Processing Pipeline (top), Rendering System (centre), and Storage Layer
(bottom). A math-content image is routed through the Mathpix API; a thumbnail is stored
directly. The format handler dispatches to the appropriate renderer before the post is
written to the database.}
\label{fig:arch}
\end{figure}
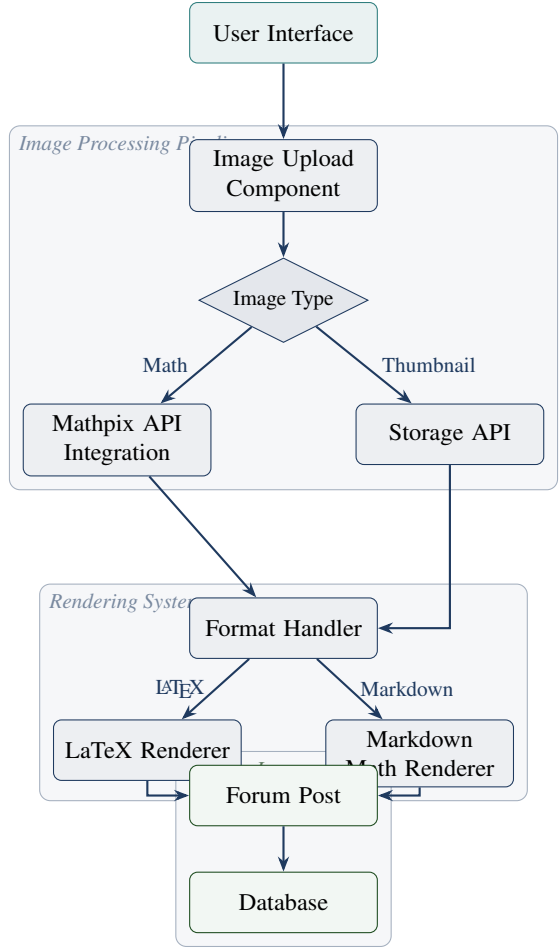

At the highest level, the system is a forum posting interface extended with an image-to-\LaTeX{}
pipeline. When a user creates a post, they can optionally upload an image containing a
mathematical expression. The system processes the image, recovers the mathematical content as
\LaTeX{} or delimited text, renders a live preview, and embeds the result in the post body.
Without the image upload, the interface functions as a standard text editor supporting
Markdown and manually entered \LaTeX{}. The full end-to-end architecture, illustrated in
Figure~\ref{fig:arch}, is organized in three layers: an \emph{Image Processing Pipeline}, a
\emph{Rendering System}, and a \emph{Storage Layer}. These are described in detail in
Sections~\ref{sec:arch}.

\section{System Architecture}
\label{sec:arch}

\subsection{Image Processing Pipeline}
The image processing pipeline is triggered when a user attaches an image to a post. The
\emph{Image Upload Component} receives the file via the browser File~API, creates a local
object URL for immediate preview, and dispatches the file to the conversion sub-system. A
lightweight type classifier then determines whether the image requires full OCR processing
(math-content path) or can be stored directly as a thumbnail (binary-content path). Images
classified as containing mathematical content are forwarded to the API Integration Layer;
thumbnails are written immediately to the Storage~API and their URLs are inserted into the
post body.

\subsection{API Integration Layer}
The API integration layer wraps the Mathpix~\cite{mathpix} OCR endpoint. The file is encoded
as multipart form data and posted to the \texttt{/v3/text} endpoint with appropriate
authentication headers. The endpoint returns a JSON object whose \texttt{latex\_styled} field
contains a pure-\LaTeX{} rendering of the expression when the model is confident, and whose
\texttt{text} field contains a plain-text version using the alternative
\texttt{\textbackslash(} \ldots \texttt{\textbackslash)} inline-math delimiters when it is
not. The integration layer extracts whichever field is populated, passes the string to the
format processing component, and surfaces errors to the user interface for retry or manual
override.

\subsection{Format Processing System}
\label{sec:format}
The format processing component is the semantic core of the pipeline. It solves a subtle but
important problem: the Mathpix API can return content in two distinct notational conventions,
and downstream renderers expect one of two others. Specifically, MathJax and KaTeX accept
\texttt{\$...\$} for inline math and \texttt{\$\$...\$\$} for display math, while the Mathpix
text output uses \texttt{\textbackslash(}, \texttt{\textbackslash)},
\texttt{\textbackslash[}, and \texttt{\textbackslash]} for the same purposes.

\begin{figure}[t]
\centering
\includegraphics[width=0.65\linewidth]{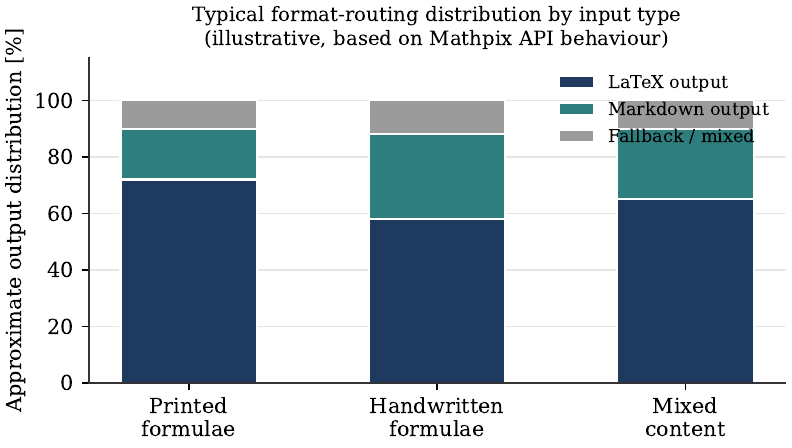}
\caption{Illustrative format-routing distribution for different input categories, based on
typical Mathpix API behaviour reported in~\cite{mathpix, pdf2latex}. Pure \LaTeX{} output
predominates for printed formulae; Markdown-delimited output is more common for handwritten
content.}
\label{fig:formatdist}
\end{figure}

The component applies the following normalisation rules when the input originates from the
\texttt{text} field rather than the \texttt{latex\_styled} field:
\begin{equation}
\texttt{\textbackslash(}\ldots\texttt{\textbackslash)} \;\longrightarrow\; \texttt{\$}\ldots\texttt{\$},
\qquad
\texttt{\textbackslash[}\ldots\texttt{\textbackslash]} \;\longrightarrow\; \texttt{\$\$}\ldots\texttt{\$\$}.
\label{eq:delim}
\end{equation}
If the input originates from \texttt{latex\_styled}, no normalisation is required and the
string is passed directly to the \LaTeX{} renderer. Figure~\ref{fig:formatdist} illustrates
the typical split of output types across different image categories.

\subsection{Rendering System}
The rendering system exposes a \emph{Format Handler} that selects between two rendering paths
based on the output of the format processing component. The \emph{LaTeX Renderer} invokes
MathJax~\cite{mathjax} or KaTeX~\cite{katex} on the processed string and injects the result
into the post composer as a rendered preview. The \emph{Markdown Math Renderer} applies a
Markdown parser with math-aware extensions, treating \texttt{\$...\$} spans as inline math
blocks before passing the result to the same MathJax/KaTeX layer. A toggle control in the
interface allows the author to switch between the two modes if the automatic selection is
incorrect. Crucially, both paths produce a live preview that updates as the user edits the
post, closing the feedback loop that forces users of manual-entry systems to submit and reload
before seeing their formula rendered.

\subsection{Storage Layer}
On submission, the post content---comprising the plain-text body, the recovered \LaTeX{}
string, the original image path, and metadata---is committed to the database according to the
schema in Table~\ref{tab:schema}. The \texttt{latex} and \texttt{latex\_Text} fields provide
redundancy between the styled and text-form outputs, enabling the application to re-render
with either convention on retrieval. The \texttt{imagePath} field retains the original image,
which is valuable for audit, re-processing with future models, and accessibility descriptions.

\begin{table}[t]
\centering
\caption{PostSchema: the database record committed on post submission.}
\label{tab:schema}
\small
\begin{tabular}{@{}lll@{}}
\toprule
\textbf{Field} & \textbf{Type} & \textbf{Purpose}\\
\midrule
\texttt{title}      & String   & Post heading.\\
\texttt{content}    & String   & Rendered post body (HTML/Markdown).\\
\texttt{latex}      & String   & \LaTeX{}-styled output from Mathpix.\\
\texttt{latex\_Text}& String   & Normalised delimiter-converted output.\\
\texttt{imagePath}  & String   & URL/path of the original uploaded image.\\
\texttt{userId}     & String   & Foreign key to the author record.\\
\texttt{timestamp}  & DateTime & UTC time of submission.\\
\bottomrule
\end{tabular}
\end{table}

\section{Implementation}
\label{sec:impl}
The system is implemented as a web application. The front end is built in React~\cite{react}
using functional components and hooks; asynchronous API calls are made with
Axios~\cite{axios}. The back end exposes a REST API and persists data to a document store
(MongoDB). MathJax v3~\cite{mathjax} handles rendering in the post composer preview; the same
engine re-renders on retrieval so that stored posts display correctly for all readers.

\paragraph{Image upload and OCR (Listing~\ref{lst:upload}).}
The \texttt{handleImageChange} handler fires when the file input changes. It creates a local
object URL for an immediate thumbnail and calls \texttt{extractLatexFromImage}, which
packages the file as \texttt{multipart/form-data} and posts it to the Mathpix
\texttt{/v3/text} endpoint. The response is passed to \texttt{processExtractedText}.

\begin{lstlisting}[caption={Image upload handler and Mathpix API call.}, label=lst:upload]
const handleImageChange = (event) => {
  const file = event.target.files[0];
  if (file) {
    const imageUrl = URL.createObjectURL(file);
    setSelectedImage(imageUrl);
    extractLatexFromImage(file);
  }
};

const extractLatexFromImage = async (file) => {
  const formData = new FormData();
  formData.append("file", file);
  const response = await axios.post(
    "https://api.mathpix.com/v3/text",
    formData,
    {
      headers: {
        "Content-Type": "multipart/form-data",
        app_id:  "[APP_ID]",
        app_key: "[APP_KEY]",
      }
    }
  );
  let extractedText =
    response.data.latex_styled || response.data.text;
  return processExtractedText(extractedText);
};
\end{lstlisting}

\paragraph{Format processing (Listing~\ref{lst:format}).}
The \texttt{processExtractedText} function checks which field was populated. If the
\texttt{latex\_styled} field was used, the string is returned unchanged. If the plain
\texttt{text} field was used, \texttt{replaceMathDelimiters} applies the substitutions
of Equation~\eqref{eq:delim}.

\begin{lstlisting}[caption={Format detection and delimiter normalisation.}, label=lst:format]
const processExtractedText = (text) => {
  if (text === response.data.text) {
    return replaceMathDelimiters(text);
  }
  return text;  // already latex_styled; no change needed
};

const replaceMathDelimiters = (text) => {
  return text
    .replace(/\\\(/g, "$")
    .replace(/\\\)/g, "$")
    .replace(/\\\[/g, "$$")
    .replace(/\\\]/g, "$$");
};
\end{lstlisting}

\section{Key Innovations}
\label{sec:innovations}
Table~\ref{tab:comparison} positions the system against representative existing tools.

\begin{table}[t]
\centering
\caption{Feature comparison with existing solutions.}
\label{tab:comparison}
\small
\setlength{\tabcolsep}{4pt}
\begin{tabular}{@{}lccccc@{}}
\toprule
\textbf{Feature}
  & \textbf{Manual \LaTeX{}}
  & \textbf{Mathpix Snip}
  & \textbf{Math SE}
  & \textbf{pix2tex}
  & \textbf{This system}\\
\midrule
Image-to-formula OCR           & --  & \checkmark & --  & \checkmark & \checkmark \\
Forum-native integration       & \checkmark & --  & \checkmark & --  & \checkmark \\
No platform switching          & \checkmark & --  & \checkmark & --  & \checkmark \\
Dual-format support            & --  & --  & --  & --  & \checkmark \\
Real-time rendered preview     & --  & \checkmark & --  & \checkmark & \checkmark \\
Mobile image capture           & --  & \checkmark & --  & --  & \checkmark \\
Automatic delimiter correction & --  & --  & --  & --  & \checkmark \\
\bottomrule
\end{tabular}
\end{table}

\paragraph{Unified Content Management.} Unlike standalone OCR tools, the system manages the
complete lifecycle---image capture, OCR, format conversion, preview, and database persistence---
as a single, stateful workflow. No clipboard, no context switching, no re-pasting. This is
especially important on mobile, where switching between applications and preserving clipboard
content is unreliable.

\paragraph{Automatic Format Detection and Delimiter Normalisation.} The format handler
transparently bridges the gap between the Mathpix API's output conventions and the
\texttt{\$}-delimited convention expected by MathJax and Markdown math extensions. This is a
latent source of errors in manual workflows: a user who pastes Mathpix output into a Markdown
editor often sees raw text rather than a rendered formula because the delimiters do not match.
The system eliminates this failure mode entirely.

\paragraph{Dual-Format Rendering with Live Preview.} Supporting both \LaTeX{} and Markdown
math in the same interface, with the appropriate engine selected automatically but overrideable
by the user, accommodates the mixed authoring conventions of real communities. The live preview
closes the compose-submit-reload loop that otherwise discourages iterative editing of
mathematical content.

\paragraph{Persistent Image Store.} Retaining the original image alongside the derived \LaTeX{}
enables re-processing with improved models in the future, provides a ground-truth reference
for error detection, and supports accessibility pipelines that generate image descriptions.

\section{The Platform as a Source of Mathematical Training Data}
\label{sec:dataset}

\subsection{The data-scarcity problem in mathematical AI}
Training AI systems to solve mathematical problems accurately requires large quantities of
high-quality, labelled data. Curating such data manually is expensive and slow: the GSM8K
dataset~\cite{gsm8k}---8.5K grade-school-level word problems with step-by-step solutions,
assembled by human writers at OpenAI---took substantial expert effort and still represents
only a narrow slice of the mathematical landscape. The MATH dataset~\cite{math_dataset}
covers competition mathematics (12,500 problems across seven subjects) but required
professional problem setters and careful difficulty tagging. Both benchmarks are invaluable,
yet static: they cannot grow as the frontier of AI capability advances or as new problem
types emerge.

Community-driven platforms such as Mathematics Stack Exchange partially address this gap by
producing large numbers of user-contributed problems and solutions, but harvesting them
requires scraping, cleaning, and converting content that was never stored in a
machine-readable format. Because mathematical expressions are entered as raw \LaTeX{} text
in unstructured prose, extracting clean problem--solution pairs demands significant
post-processing and still loses the original handwritten or typeset images. The proposed
system eliminates this bottleneck by design.

\subsection{How the platform generates training data}
Every post on the platform is already a structured data record. A user uploads an image of a
problem---handwritten, typeset, or photographed from a textbook---which the system converts
to \LaTeX{} and stores alongside the original image in the PostSchema
(Table~\ref{tab:schema}). Other users then post solutions as replies, entering their working
via the same image-to-\LaTeX{} pipeline or by typing directly. The result is a collection
of tuples of the form:
\begin{equation}
(\underbrace{I_{\mathrm{prob}}}_{\text{problem image}},\;
 \underbrace{L_{\mathrm{prob}}}_{\text{problem in \LaTeX{}}},\;
 \underbrace{S_1, S_2, \ldots}_{\text{solutions}},\;
 \underbrace{v_1, v_2, \ldots}_{\text{community votes}})
\label{eq:tuple}
\end{equation}
where each solution $S_i$ is itself a structured document containing natural-language
explanation, \LaTeX{} expressions, and intermediate steps. This is precisely the
\emph{chain-of-thought} format~\cite{wei2022} that has proven most effective for training
language models to reason about mathematics: rather than recording only a final answer, the
platform naturally captures the full reasoning trace that a human solver writes out.

\subsection{Data quality and scale}
Several properties of the platform contribute to data quality at scale.

\paragraph{Community verification.} Votes and accepted-answer flags provide a continuous,
crowd-sourced quality signal. A solution that receives many upvotes and is accepted by the
question author is strong evidence of correctness---a signal that is absent from synthetically
generated datasets and difficult to replicate without a live community.

\paragraph{Multiple solutions per problem.} A single problem typically attracts several
independent solutions employing different methods (e.g.\ algebraic, geometric, and
algorithmic approaches to the same question). This diversity is valuable for training models
that are robust across solution strategies and for evaluating whether a model understands
\emph{why} an answer is correct rather than merely memorising a pattern.

\paragraph{Natural difficulty stratification.} Forum tags and categories (e.g.\ calculus,
linear algebra, number theory, competition problems) provide automatic difficulty and
subject-area labels without additional annotation cost. A model fine-tuned on data from this
platform therefore comes with built-in metadata for curriculum-style training.

\paragraph{Persistent image--\LaTeX{} alignment.} The \texttt{imagePath} field retains the
original photograph of the problem, giving downstream researchers access to both the visual
and the symbolic representation. This is useful for training multimodal models that read
problems directly from images---the setting that most closely mirrors how a student
encounters mathematics in the real world.

\subsection{Implications for AI training}
Table~\ref{tab:dataset} compares the platform's data-generation properties against existing
mathematical datasets.

\begin{table}[t]
\centering
\caption{Comparison of mathematical dataset sources. ``Continuous'' means the dataset grows
automatically with platform usage without additional human curation cost.}
\label{tab:dataset}
\small
\begin{tabular}{@{}lcccccc@{}}
\toprule
\textbf{Source}
  & \textbf{Scale}
  & \textbf{Continuous}
  & \textbf{Image}
  & \textbf{Steps}
  & \textbf{Verified}
  & \textbf{Multi-solution}\\
\midrule
GSM8K~\cite{gsm8k}        & 8.5K  & --  & --  & \checkmark & expert  & -- \\
MATH~\cite{math_dataset}  & 12.5K & --  & --  & \checkmark & expert  & -- \\
Math Stack Exchange       & $>$10M & \checkmark & --  & \checkmark & crowd  & \checkmark \\
This platform             & grows & \checkmark & \checkmark & \checkmark & crowd  & \checkmark \\
\bottomrule
\end{tabular}
\end{table}

The platform's most significant advantage over Math Stack Exchange is the persistent image
store and the machine-readable \LaTeX{} alignment guaranteed by the conversion pipeline.
Every problem arrives with a photograph of the original source---a textbook page,
handwritten homework, or printed exam---making the data suitable for training vision-language
models of the kind now central to mathematical AI research. Researchers wishing to use the
accumulated data for training need only query the PostSchema database: the problem image,
the \LaTeX{} representation, and the community-validated solutions are available in a single,
consistently structured record, with no scraping or re-parsing required.

\section{Limitations and Future Work}
\label{sec:limits}
The system currently relies on the Mathpix API as its sole OCR backend. This introduces a
dependency on an external commercial service, with associated costs, latency, and availability
risks. Future work includes integration of open-source alternatives such as
pix2tex~\cite{pix2tex}, which would allow on-premise deployment and eliminate per-call fees.

Mathpix's performance is strong on standard printed and handwritten formulae but degrades on
complex multi-line derivations, chemistry structures, commutative diagrams, and other
non-standard mathematical notation. Extending the system to handle these categories---either
through supplementary models or through a structured editor for the cases OCR cannot resolve---
is a natural next step.

The format-detection heuristic of Section~\ref{sec:format} uses a simple string comparison
(checking whether the extracted text matches \texttt{response.data.text}) to decide whether
delimiter normalisation is required. A more robust classifier that inspects the string content
directly would be more resilient to API changes. Additionally, the current implementation does
not validate the syntactic correctness of the recovered \LaTeX{} before rendering; integrating
a lightweight \LaTeX{} parser for pre-flight error detection would improve the user experience
for malformed or partially-recognised expressions.

Finally, the system has been evaluated informally; a user study measuring posting time,
error rate, and user satisfaction compared against manual-entry baselines is important future
work to quantify the gains suggested by the step-count reduction of Figure~\ref{fig:steps}.

\section{Conclusion}
\label{sec:conclusion}
We have presented a mathematical forum system that integrates image-to-\LaTeX{} conversion
directly into the post-creation workflow, reducing the number of user interaction steps from
four or five to one. The architecture separates concerns cleanly across an image processing
pipeline, a rendering system, and a storage layer, and introduces a format-handler component
that automatically bridges the delimiter mismatch between the Mathpix API and web rendering
engines. The system supports both desktop and mobile clients, provides a live preview of
rendered mathematics before submission, and retains the original image for future
re-processing. A US provisional patent application (No.~63/727,195) has been filed.

The system has a second, compounding value: every problem posted and every solution
contributed is stored as a structured, machine-readable record---image, \LaTeX{},
natural-language reasoning, and community quality signal---in the format most useful for
training AI systems to solve mathematics accurately. As the platform grows, it becomes
simultaneously more useful to its users and more valuable as a training resource, creating
a virtuous cycle between human collaboration and AI capability. The scarcity of large,
high-quality, verified mathematical datasets~\cite{gsm8k,math_dataset} makes this
by-product of ordinary forum activity a meaningful contribution in its own right.

We anticipate that integrating open-source OCR backends, conducting controlled user studies,
and releasing a curated snapshot of the accumulated problem--solution dataset will be the
most fruitful directions for follow-on work.

\section*{Acknowledgements}
The authors thank the open-source communities behind MathJax, KaTeX, and React for the tools
on which this system is built.

\end{document}